\documentclass[11pt]{article}
\usepackage{coling2018}
\usepackage{times}
\usepackage{url}
\usepackage{latexsym}
\usepackage{times}
\usepackage{epsfig}
\usepackage{graphicx}
\usepackage{amsmath}
\usepackage{amssymb}
\usepackage{caption}
\usepackage{subcaption}
\usepackage{multirow}
\usepackage{color}

\newcommand{\etal}{et al.}

\title{iParaphrasing: Extracting Visually Grounded Paraphrases via an Image}

\author{Chenhui Chu,$^1$ Mayu Otani$^2$ and Yuta Nakashima$^1$\\
  $^1$Institute for Datability Science, Osaka University \\
  $^2$CyberAgent, Inc. \\
{\tt chu,n-yuta@ids.osaka-u.ac.jp, otani\_mayu@cyberagent.co.jp}} 

\date{}

\begin{document}
\maketitle
\begin{abstract}
A paraphrase is a restatement of the meaning of a text in other words. Paraphrases have been studied to enhance the performance of many natural language processing tasks. 
In this paper, we propose a novel task {\it iParaphrasing} to extract {\it visually grounded paraphrases (VGPs)}, which are different phrasal expressions describing the same visual concept in an image. These extracted VGPs have the potential to improve language and image multimodal tasks such as visual question answering and image captioning. How to model the similarity between VGPs is the key of iParaphrasing. We apply various existing methods as well as propose a novel neural network-based method with image attention, and report the results of the first attempt toward iParaphrasing.
\end{abstract}
 
\section{Introduction}
\begin{figure}[t]
\centering{\includegraphics[width=\hsize]{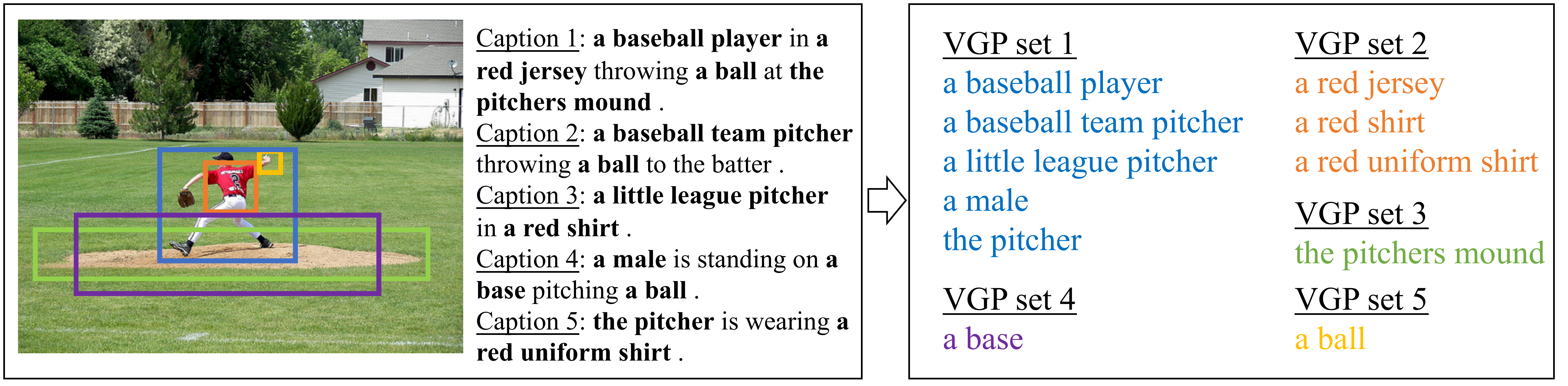}} 
\caption{An example from the Flickr30k entities dataset, in which an image is described by five captions (entities in the captions are marked in bold). Our task is to extract the entities that describe the same visual concept (represented as an image region) in the image as VGPs. Note that the image regions are not given as input but are drawn here for comprehensibility.}
\label{fig:iparaphrase}
\end{figure}

%
\blfootnote{
 \hspace{-0.65cm}  
This work is licensed under a Creative Commons Attribution 4.0 International License. License details: \url{http://creativecommons.org/licenses/by/4.0/}
}
A paraphrase is a restatement of the meaning of a word, phrase, or sentence within the
context of a specific language (e.g., ``a red jersey'' and ``a red uniform shirt'' in Figure \ref{fig:iparaphrase} are paraphrases) \cite{Rahul_CL_2013}. Paraphrases have been exploited for natural language understanding, and shown to be very effective for various natural language processing (NLP) tasks, including question answering \cite{riezler-EtAl:2007:ACLMain}, summarization \cite{zhou-EtAl:2006:HLT-NAACL06-Main2}, machine translation \cite{Chu_LREC_2016}, text normalization \cite{ling-EtAl:2013:EMNLP}, textual entailment recognition \cite{Androutsopoulos:2010:SPT:1892211.1892215}, and semantic parsing \cite{berant-liang:2014:P14-1}. 

In this paper, we propose a novel task named, {\it iParaphrasing}, to extract {\it visually grounded paraphrases (VGPs)}. We define VGPs as different phrasal expressions that describe the same visual concept in an image. Nowadays, with the spread of the web and social media, it is easy to collect large amounts of images with their describing text. For example, different news sites release news with the same topic using the same image; photos with many comments are posted to social networking sites and blogs. As these describing texts are written by different people but about the same image, there are potentially large amounts of VGPs in the describing text (Figure 1). We aim to accurately extract these paraphrases using the image as a pivot to associate different phrases. 

The extracted VGPs can be applied to various computer vision (CV) and NLP tasks, such as image captioning \cite{Vinyals_2015_CVPR} and visual question answering (VQA) \cite{Wu:2017:CVIU}, for the better understanding of both images and languages. For example, a VQA system must understand queries of different expressions about the same visual concept (e.g., ``a male'' and ``the pitcher'' in Figure \ref{fig:iparaphrase}) in order to answer a question properly. VGPs can also be applied to
the evaluation of image captioning systems 
in the similar way as paraphrases have been applied for machine translation evaluation \cite{Snover:2009:TPS:1743627.1743646}. 

As a pioneering study, we work on iParaphrasing on the Flickr30k entities dataset \cite{Plummer_2015_ICCV}. This dataset contains 30k images with 5 captions per image annotated via crowdsourcing, which can be seen as a very small subset of the data available in the web and social media. Figure 1 shows an example image together with its five captions taken from this dataset. 
In the Flickr30k entities dataset, entities (i.e., noun phrases) in the captions have been manually aligned to their corresponding image regions \cite{Plummer_2015_ICCV}. Therefore, we can obtain a set of phrases annotated with the same image region. This set of phrases are used as the ground truth VGPs in our study. The goal of this work is to extract these VGPs.


We formulate our task as a clustering task (Section \ref{sec:extraction}), 
where the similarity between each entity pair is crucial for the performance. We apply many different unsupervised similarity computation methods (Section \ref{sec:unsupervised}) including phrase localization-based similarity \cite{Plummer_2017_ICCV} (Section \ref{sec:phrase_localization}), translation probability-based similarity \cite{koehn-EtAl:2007:PosterDemo} (Section \ref{sec:translation_probability}), and embedding-based similarity \cite{DBLP:journals/corr/abs-1301-3781,DBLP:journals/corr/KleinLSW14,Plummer_2015_ICCV}  (Section \ref{sec:embedding}). In addition, we propose a supervised neural network (NN)-based  method using both textual and visual features to explicitly model the similarity of an entity pair as VGPs (Section \ref{sec:nn}). Experiments show that our proposed NN-based method outperforms the other methods.\footnote{Codes and data for reproducing the results reported in this paper are available at \url{https://github.com/ids-cv/coling_iparaphrasing}}



\section{Related Work}
\subsection{Paraphrase Extraction}


Previous studies extract paraphrases from either monolingual corpora or bilingual parallel corpora. One major approach is to use the distributional similarity \cite{harris54} with regular monolingual corpora (a large collection of text in a single language) \cite{Lin:2001:DSI:502512.502559,bhagat-ravichandran:2008:ACLMain,marton-callisonburch-resnik:2009:EMNLP}, or monolingual comparable corpora (a set of monolingual corpora that describe roughly the same
topic in the same language) \cite{Barzilay:2006:HLT-NAACL03,chen-dolan:2011:ACL-HLT2011}. Distributional similarity stems from the distributional hypothesis \cite{harris54}, stating that words/phrases that share similar meanings should appear in similar distributions. This approach sometimes suffers from noisy results, because the distributed similarity often maps antonyms to closer points. Some methods try to extract paraphrases from monolingual parallel corpora (a collection of sentence level paraphrases) \cite{arase-tsujii:2017:EMNLP2017,maccartney-galley-manning:2008:EMNLP}, but such monolingual parallel corpora are rarely available. 

Bilingual parallel corpora (a collection of sentence-aligned bilingual text) enjoys more availability than monolingual parallel corpora as they are mandatory for training machine translation systems.  Bilingual parallel corpora can be used for paraphrase extraction, with bilingual pivoting \cite{bannard-callisonburch:2005:ACL}. This method assumes that two source phrases are a paraphrase pair if they are translated to the same target phrase. Bilingual pivoting has been further refined by using syntax information \cite{callisonburch:2008:EMNLP} or mutual information \cite{Kajiwara:2017:IJCNLP}. These methods have led to the construction of a multilingual paraphrase database \cite{GANITKEVITCH14.659.L14-1520}. 

Note that our definition of paraphrases may look different from the studies mentioned above, as our paraphrases are a set of noun phrases that represent the same visual concept. Our idea to extract paraphrases under this definition is to use image captioning datasets \cite{Young:TACL:2014,DBLP:journals/corr/LinMBHPRDZ14}, which usually contain several captions for each image, and currently scale to sub-million images, instead of a bilingual parallel corpus with limited availability. To the best of our knowledge, this is the first study that aims to extract paraphrases from such multimodal datasets consisting of images and their captions.\footnote{Although \cite{Plummer_2015_ICCV} annotated the VGPs in the Filickr30k entities dataset, they did not propose any methods to extract them. \cite{Regneri:TACL2013} collected sentence level paraphrases by aligning video scripts with the same time  frame; these sentence level paraphrases are essentially similar to captions of an image.}



\subsection{Coreference Resolution}
Coreference resolution is a task to find the expressions that refer to the same entity in a text \cite{Soon:2001:MLA:972597.972602,lee-EtAl:2017:EMNLP2017}. Our task in this paper focuses on extracting entities that describe the same visual concept, making the formulation similar to coreference resolution. Our task differs from conventional coreference resolution that it requires visual grounding. In addition, the targets of coreference resolution are the entities in a sentence or a document, while our targets are the entities in the captions of an image that are quasi-paraphrases but are not related to each other in discourse level like sentences in a document. 
For coreference resolution, the context in a sentence or discourse information in a document are crucial, but discourse information does not exist in our task.\footnote{If we treat multiple captions as a document forcibly, we could apply a coreference resolution approach for our task.} 
In the context of vision and language tasks, Kong\etal~\shortcite{Kong_2014_CVPR} used noun/pronoun coreference resolution in sentential descriptions of RGB-D scenes for improving 3D semantic parsing. The texts being handled in their work are either sentences or documents, and their targets are limited to noun words and pronouns but we extract noun phrases.
Because our goal is not limited to entities but arbitrary phrases, we believe that comparing to coreference resolution, iParaphrasing is a more forethoughtful name to define our task for future research.


\subsection{Phrase Localization}
Phrase localization is a task to find an image region that corresponds to a given phrase in a caption, which is closely related to our VGP extraction task. Plummer
\etal~\shortcite{Plummer_2015_ICCV} pioneered this work, in which they annotated phrase-region alignment in the Flickr30k image-caption dataset \cite{Young:TACL:2014} and released it as the Flickr30k entities dataset. They also proposed a method based on canonical correlation analysis (CCA) \cite{Hardoon:2004:CCA:1119696.1119703} that learns joint embeddings of phrases and image regions for associating them. Wang \etal~\shortcite{Wang_2016_CVPR} proposed joint embeddings using a two-branch NN. Fukui \etal~\shortcite{fukui-EtAl:2016:EMNLP2016} used a multimodal compact bilinear pooling method to combine textual and visual embeddings. Rohrbach \etal~\shortcite{Rohrbach_2016_ECCV} proposed a convolutional NN (CNN)-recurrent NN (RNN)-based method for this task. They learn to detect a region for a given phrase and then reconstruct the phrase using the detected region. Wang \etal~\shortcite{Wang_2016_ECCV} noticed that the relationships between phrases should agree with their corresponding regions, and proposed a joint matching method, but their method only considers the ``has-a'' relationship that is explicitly indicated by possessive pronouns. 
Previous studies rely on region proposal to produce a number of region candidates for phrase localization, Yeh \etal~\shortcite{DBLP:conf/nips/YehXHDS17} proposed a unified framework that can search over all possible regions.
Plummer \etal~\shortcite{Plummer_2017_ICCV} used 
spatial relationships between pairs of entities connected by verbs or prepositions, which achieved the state-of-the-art performance. 
In this paper, we use the current state-of-the-art phrase localization method of \cite{Plummer_2017_ICCV} as a baseline for VGP extraction.

\subsection{Other Vision and Language Tasks}

Vision and language tasks have been a hot research area recently in both the CV and NLP communities. Various  efforts have been made for many multimodal tasks such as image object/region referring expression grounding \cite{D14-1086,Mao_2016_CVPR,Hu_2017_CVPR,Cirik_AAAI2018} 
visual captioning \cite{Vinyals_2015_CVPR,pmlr-v37-xuc15,Bernardi:2016:JAIR,laokulrat-EtAl:2016:COLING}, text-image retrieval \cite{Otani_2016_ECCV}, visual question answering \cite{Wu:2017:CVIU}, visual dialog \cite{visdial,visdial_rl} and video event detection \cite{phan-EtAl:2016:COLING}. Some researchers also have employed images for improving NLP tasks, such as multimodal machine translation \cite{specia-EtAl:2016:WMT}, cross-lingual document retrieval \cite{funaki-nakayama:2015:EMNLP}, and textual entailment recognition \cite{han-martinezgomez-mineshima:2017:EMNLP2017}. iParaphrasing is a novel CV$+$NLP task, which to the best of our knowledge has not been studied before and can boost the performance of various multimodal and NLP tasks.


\section{Paraphrase Extraction via Clustering}
\label{sec:extraction}
We formulate the paraphrase extraction from the Flickr30k entities dataset as a clustering task. Given an image and all the entities in the corresponding captions, the task is to cluster the entities\footnote{In this paper, we assume that entities are given. In the case that entities are not given, we can easily extract them by chunking the noun phrases.} to its corresponding visual concepts represented as image regions. 
The number of clusters (i.e., the number of paraphrase sets in a set of an image and captions) is not explicitly given in our task.
Therefore, we apply the affinity propagation algorithm \cite{Frey2008} to cluster entities, which can estimate the number of clusters as well.

Affinity propagation creates clusters by iteratively sending two types of messages between pairs of entities until convergence. The first type is the responsibility $r(i, j)$ sent from entity $i$ to candidate representative entity $j$, indicating the strength that entity $j$ should be the representative entity for entity $i$, which is defined as:
\begin{equation}
r(i,j) \leftarrow s(i,j)-\max_{\forall j^\prime\neq j}\{a(i,j^\prime)+s(i,j^\prime)\}
\end{equation}
where $s(i,j)$ is the similarity between entities $i$ and $j$.
The second type is the availability $a(i, j)$ sent from candidate representative entity $j$ to entity $i$, 
indicating to what degree that candidate representative entity $j$ is the cluster center for entity $i$, which is defined as:
\begin{equation}
a(i,j) \leftarrow \min\Bigl\{0, r(j,j)+\sum_{\forall i^\prime \not\in \{i,j\}}{\max\{0,r(i^\prime, j)\}\Bigr\}}
\end{equation}
At the beginning, the values of $r(i,j)$ and $a(i,j)$ are set to zero, and they are updated in every iteration until convergence. We optimize the number of clusters on a validation split by adjusting the preference (i.e., self similarity $s(i,i)$) of affinity propagation.

\begin{figure}[t]
\centering{\includegraphics[width=0.35\hsize]{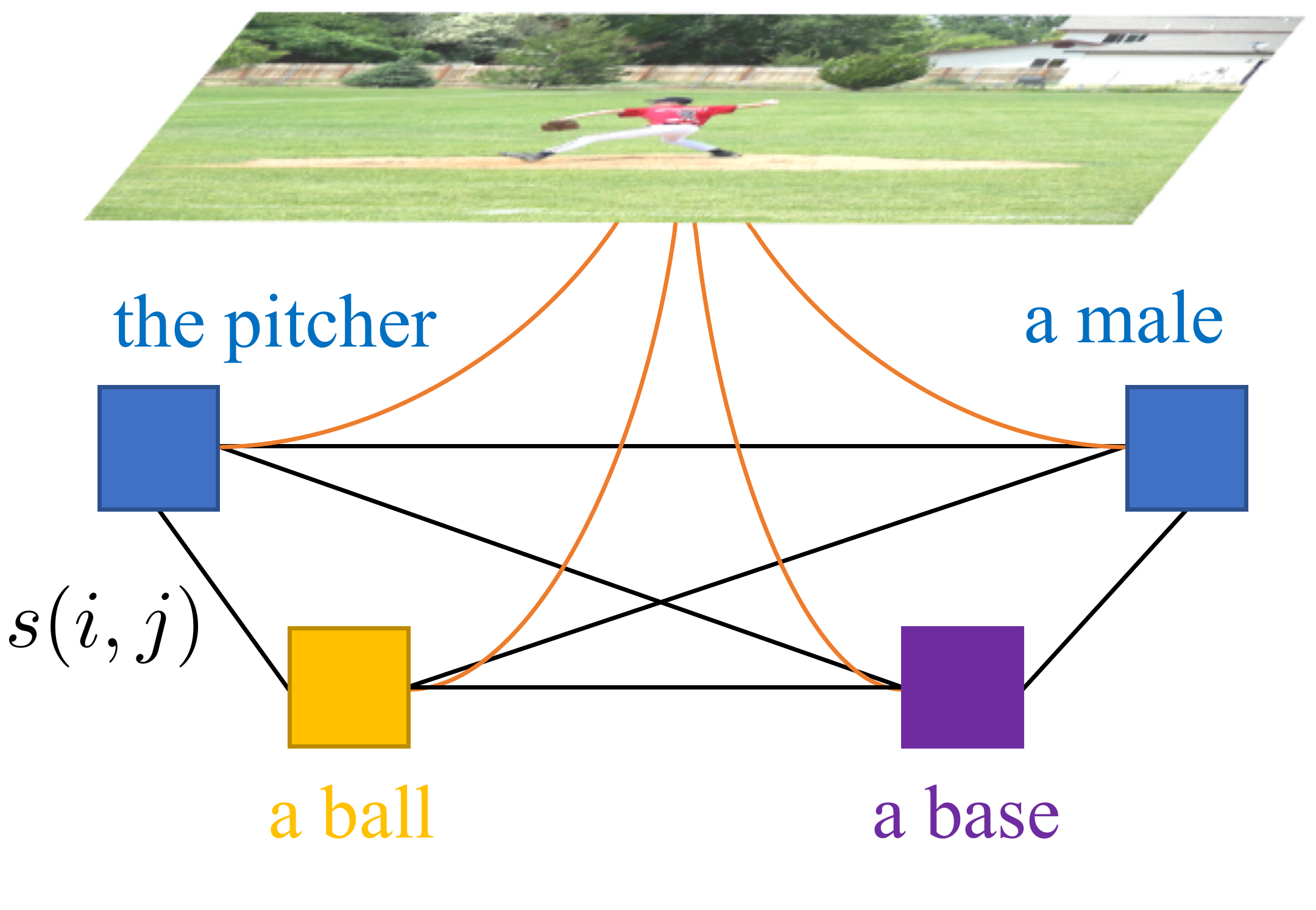}} 
\caption{An overview of our VGP extraction formulation. We extract VGP via clustering, where the entity-entity similarity $s(i,j)$ is the key. We compare both unsupervised and supervised methods using entity-image and entity-entity associations for computing this similarity.}
\label{fig:overview}
\end{figure}

Figure \ref{fig:overview} shows an overview of our formulation, where
the similarity between the entities is the key. We apply various unsupervised methods for computing this similarity, and propose a supervised NN-based model. 


\section{Unsupervised Similarity Methods}
\label{sec:unsupervised}
We apply phrase localization for modeling the entity-entity similarity based on entity-image association (Section \ref{sec:phrase_localization}). In addition, we apply various methods for modeling the  entity-entity similarity directly (Sections \ref{sec:translation_probability},  and \ref{sec:embedding}). 

\subsection{Phrase Localization-Based Similarity}
\label{sec:phrase_localization}
The similarity between entities $i$ and $j$ is defined as:
\begin{equation}
\label{eq:pl}
s(i,j)=\sum_{r_m\in{R}}{p(i|r_m)p(j|r_m)}
\end{equation}
where $R$ is a set of image regions that are aligned to both entities $i$ and $j$ obtained with the phrase localization method of \cite{Plummer_2017_ICCV};
$p(i|r_m)$ is the localization probability of $r_m$ for $i$, defined as:
\begin{equation}
p(i|r_m) = \frac{l(i,r_m)}{\sum_{r_m\in{R}}l(i,r_m)} 
\end{equation}
where $l(i,r_m)$ is the localization score of region $r_m$ for entity $i$ obtained using the method of \cite{Plummer_2017_ICCV}.

\subsection{Translation Probability-Based Similarity}
\label{sec:translation_probability}
The similarity between entities $i$ and $j$ is defined as:
\begin{equation}
s(i,j)=p(i|j)p(j|i)
\end{equation}
where $p(i|j)$ and $p(j|i)$ are the direct and inverse translation probabilities of an entity pair $i$ and $j$, which are calculated using a conventional statistical machine translation (SMT) \cite{koehn-EtAl:2007:PosterDemo} method:
\begin{enumerate}
\item Generate a pseudo parallel corpus using the captions in the dataset, which treats the 5 captions for each image as monolingual parallel sentences and pair each of the sentences that leads to $\binom{5}{2}=10$ sentence pairs per image.
\item Apply word alignment to the parallel corpus using IBM alignment models \cite{brown-EtAl:1993} in two directions with the grow-diag-final-and heuristic \cite{koehn-EtAl:2007:PosterDemo} to align the words in each caption pair. 
\item From the word-aligned parallel corpus,  extract entity pairs such that the words inside an entity pair are aligned. Then $p(i|j)$ and $p(j|i)$ are calculated as follows:
\begin{equation}
p(i|j)=\frac{c(i,j)}{\sum_{k}{c(i,k)}}, \quad
p(j|i)=\frac{c(i,j)}{\sum_{k}{c(j,k)}}
\end{equation}
where $c(i,j)$ is the number of co-occurrence of $i$ and $j$ in the word-aligned corpus.
\end{enumerate}

\subsection{Embedding-Based Similarity}
\label{sec:embedding}
In this method, the similarity between entities $i$ and $j$ is defined as:
\begin{equation}
s(i, j)=\frac{\mathbf{t}_i^\top \mathbf{t}_j}{\|\mathbf{t}_i\| \| \mathbf{t}_j\|}
\end{equation}
where $\mathbf{t}_i$ and $\mathbf{t}_j$ are the phrase embeddings of $i$ and $j$.
We compare three different methods for phrase embeddings.
\subsubsection{Word Embedding Average}
\label{sec:avg}
We represent each word with a 300 dimensional word2vec \cite{DBLP:journals/corr/abs-1301-3781} vector pre-trained on the Google News corpus.\footnote{https://github.com/mmihaltz/word2vec-GoogleNews-vectors} We remove stop words in each entity, and calculate the representation of each entity using the average of all word embeddings.

\subsubsection{Fisher Vector}
\label{sec:fv}
Fisher vector is a pooling over word2vec vectors of individual words \cite{DBLP:journals/corr/KleinLSW14}, which has been used in the phrase localization task for representing the entities \cite{Plummer_2015_ICCV}.
To compute the Fisher vector for an entity,  we represent the entity by the HGLMM Fisher vector encoding \cite{DBLP:journals/corr/KleinLSW14} of the word vectors, following \cite{Plummer_2015_ICCV}.\footnote{The Fisher vector is constructed with 30 centers of both first and second order information, which results in a very sparse vector whose dimensionality is $300 \times 30 \times 2 = 18000$. Therefore, we apply principal component analysis (PCA) to convert it to a lower dimensionality of 4,096.}

\subsubsection{Fisher Vector with CCA}
\label{sec:fv_cca}
Projecting the feature vectors of image regions and entities to a shared semantic space can provide strong associations between the image regions and entities, which has the potential to improve the performance of VGP extraction. 
Therefore, we learn a CCA projection on the Flickr30k entities dataset for the image region feature vectors and entity feature vectors with \cite{Plummer_2015_ICCV}, in which the normalized CCA formulation of \cite{Gong:2014:MES:2584252.2584265} is used. The columns of the CCA projection matrices are scaled by the eigenvalues, and the feature vectors are projected by these matrices and normalized to the dimensionality of 4,096. The image region feature vectors are extracted using Faster R-CNN \cite{NIPS2015_5638}.\footnote{https://github.com/ShaoqingRen/faster\_rcnn} We use Fisher vectors for entity feature vectors.

\section{Supervised Similarity Model Based on Neural Network with Image Attention}
\label{sec:nn}

\begin{figure*}[t!]
\centering
\includegraphics[width=\hsize]{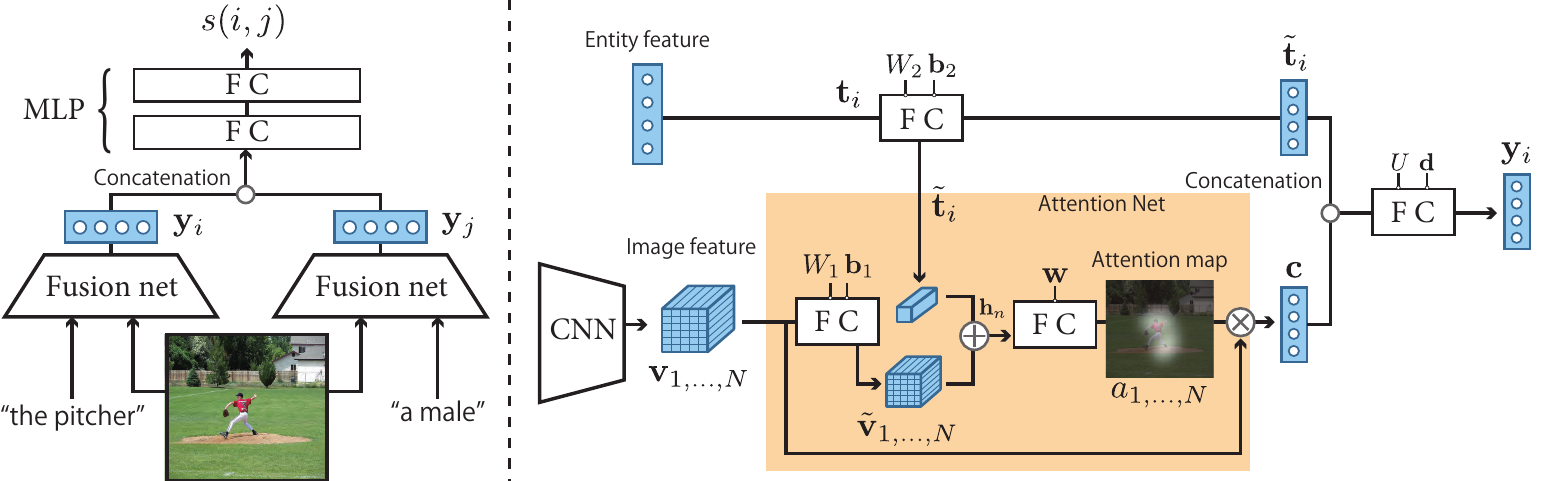}
\caption{Our supervised NN with image attention-based similarity model (left) and its fusion sub-network (right).}
\label{fig:model}
\end{figure*}

We propose a NN-based supervised model.
This model computes the similarities of entity pairs as VGPs by explicitly modeling the associations between them and an image. Figure \ref{fig:model} illustrates our proposed NN model.\footnote{\cite{TACL831} proposed a CNN network with attention for sentence level paraphrase identification; our model differs from theirs that we fuse both textual and visual information while theirs is a text only model.} Given an entity pair and its corresponding image, we construct two separated {\it fusion nets} for each entity (Figure \ref{fig:model} (right)). Note that parameters of these two fusion net are shared.
A fusion net represents an entity with a concatenation of its entity feature vector and visual context vector. The visual context vector is computed with an attention mechanism, indicating to which part of the image should be paid attention, in order to judge whether the entity pair is VGP or not. The outputs of the two fusion nets are then fed into a multilayer perceptron (MLP) to compute the similarity of the two entities.

Formally, let $V$ be a $196 \times 512$ feature map\footnote{An image is split into $14 \times 14=196$ sub-images, and represented as a $196 \times 512$ feature map.} extracted from the \texttt{conv5\_3} layer in the VGG-16 network \cite{DBLP:journals/corr/SimonyanZ14a} for an input image; 
$\mathbf{v}_n$ is a 512 dimensional vector at position \(n\) of $V$. Given an entity feature vector $\mathbf{t}_i$ and $\mathbf{v}_n$, we first transform them with fully connected (FC) layers whose unit sizes are 512:
\begin{equation}
\tilde{\mathbf{v}}_n = \mathrm{norm_{L2}} (W_1 \mathbf{v}_n + \mathbf{b}_1), \quad
\tilde{\mathbf{t}}_i = \mathrm{norm_{L2}} (W_2 \mathbf{t}_i + \mathbf{b}_2)
\end{equation}
where \(\mathrm{norm_{L2}}(\cdot)\) indicates L2 normalization to an input vector. 
We then compute an attention value $a_n$ for $\mathbf{v}_n$ as:
\begin{equation}
\mathbf{h}_n = \mathrm{relu}(\tilde{\mathbf{v}}_n + \tilde{\mathbf{t}}_i), \quad
e_n = \mathbf{w}^\top \mathbf{h}_n, \quad
a_n = \frac{\exp(e_n)}{\sum_{n=1}^{N} \exp(e_n)}
\end{equation}
where \(N = 196\). 
After obtaining $a_n$, we fuse a visual and an entity feature vector to $\mathbf{y}_i$ as:
\begin{equation}
\mathbf{c} = \sum_{n=1}^{N} a_n \mathbf{v}_n, \quad
\mathbf{y}_i = U [\mathrm{norm_{L2}}(\mathbf{c}), \tilde{\mathbf{t}}_i] + \mathbf{d}
\end{equation}
where \([\cdot, \cdot]\) indicates the concatenation of two vectors, $\mathbf{c}$ is a visual context vector.
We compute fusion feature vectors $\mathbf{y}_i$ and $\mathbf{y}_j$ with the corresponding image. Finally, we feed them to a two-layer MLP network with ReLU non-linearity, whose unit sizes are 128 and 1, respectively, to produce the similarity of the entity pair.

\section{Experiments}
\subsection{Settings}
\label{sec:settings}
We conducted experiments on the Flickr30k entities dataset \cite{Plummer_2015_ICCV}. This dataset contains 31,837 images, which is described with 5 captions annotated via crowdsourcing. We followed the 29,873 training, 1,000 validation, and 1,000 test image splits used in the phrase localization task \cite{Plummer_2015_ICCV}. 
Our task is to automatically cluster the entities in the captions that describe the same visual concept (i.e., region in the dataset) in the image as VGPs.
Entities that share the same ID and group type (e.g., ``a red jersey,'' ``a red shirt'' and ``a red uniform shirt'' in Figure \ref{fig:iparaphrase} share the same entity ID and group type ``/EN\#19026/clothing'') are treated as the ground truth VGP clusters in our evaluation.\footnote{There is 
an entity type named ``notvisual'' in the dataset (e.g., ``the batter'' in Figure \ref{fig:iparaphrase}), which means this entity has no corresponding visual regions in the image. In our evaluation, we excluded this ``notvisual'' type, because all entities that are not visual are annotated with the same entity ID and thus ground truth VGPs for these ``notvisual'' entities are unavailable in the dataset. There are entity pairs in the dataset that are the same after removing the stop words (e.g., ``a man'' and ``the man''), we treated them as one entity for evaluation. In addition, entities that do not have corresponding regions in the image were excluded from evaluation.} 
As stop words should not be considered for computing the entity similarities, we preprocessed the entities in the dataset by removing stop words for all the methods.

We evaluated both clustering and pairwise performance.\footnote{We did not report phrase localization accuracies for our proposed NN-based supervised model, because attention is different from phrase localization. Instead of determining the best region corresponding to the given phrases, attention provides attention probabilities to the 196 sub-images, which cannot be used for evaluation directly. A soft-accuracy metric could be reported for discussion, but this metric is not directly comparable to previous phrase localization studies.} The entity clustering performance for each image was measured with adjusted Rand index (ARI) \cite{Hubert1985}. We used the  implementation in the Scikit-learn machine learning toolkit \cite{scikit_learn}\footnote{http://scikit-learn.org/stable/modules/clustering.html\#adjusted-rand-index} for computing ARI.
We report the mean of ARI scores for all the images in the test split. To evaluate the performance for clustering, we optimized the number of clusters by adjusting
the preference for affinity propagation on the validation split to maximize the ARI using the Bayesian optimization algorithm \cite{mockus1989bayesian} implemented in GPyOpt.\footnote{https://github.com/SheffieldML/GPyOpt}
The pairwise performance was evaluated with precision, recall, and F-score, defined as:
\begin{equation}
\mathrm{precision} = \frac{\mathrm{\#predicted\_positive}}{\#\mathrm{predicted}},
\mathrm{recall} = \frac{\mathrm{\#predicted\_positive}}{\#\mathrm{true\_positive}},
\mathrm{F\mathchar`-score} = 2\times\frac{\mathrm{precision\times recall}}{\mathrm{precision+recall}}
\end{equation}
where an entity pair with a similarity higher than a threshold is treated as $predicted$, which is compared against the ground truth to judge whether it is $predicted\_positive$ or not.
We report the performance using the similarity threshold tuned on the validation split that maximizes the F-score.

We used the affinity propagation implementation\footnote{http://scikit-learn.org/stable/modules/generated/sklearn.cluster.AffinityPropagation.html} in Scikit-learn for clustering. We compared the performance of the different similarity methods described in Sections \ref{sec:unsupervised} and \ref{sec:nn}, where the detailed settings for the methods  were as follows:
\begin{itemize}
\item Phrase localization (PL): we used the pl-clc toolkit,\footnote{https://github.com/BryanPlummer/pl-clc} which is an implementation of the localization method of \cite{Plummer_2017_ICCV}. For $R$ in equation \ref{eq:pl}, we used the  top $30$ localization candidates for each entity. The localization scores for each entity and region pair obtained with \cite{Plummer_2017_ICCV} were used to compute the similarity.
\item Translation probability (TP): to get the entity translation probabilities, we first applied the GIZA++ toolkit\footnote{http://code.google.com/p/giza-pp} that is an implementation of the IBM alignment models \cite{brown-EtAl:1993} on the pseudo parallel corpus, and then a phrase table was extracted and the phrasal translation probabilities were calculated using state-of-the-art SMT toolkit Moses \cite{koehn-EtAl:2007:PosterDemo}.
\item Word embedding average (WEA): see the detailed setting in Section \ref{sec:avg}.
\item Fisher vector (FV): entity feature vectors were computed using the Fisher vector toolkit released by the authors,\footnote{https://owncloud.cs.tau.ac.il/index.php/s/vb7ys8Xe8J8s8vo} following the settings described in Section \ref{sec:fv}.
\item Fisher vector w/ CCA (FV+CCA): image region feature vectors and entity feature vectors were projected into a 4,096 dimensional space CCA trained on the training split of the Flickr30k entity dataset (Section \ref{sec:fv_cca}).
\item Supervised NN (SNN): to show the effectiveness of the fusion net (Section \ref{sec:nn}), we compared a supervised NN-based setting that only feeding the entity feature vectors to the MLP (Figure \ref{fig:model} (left)) for paraphrase similarity prediction. 
This setting only uses entity feature vectors as input for the NN. It was trained on the training split of the Flickr30k entity dataset. We used all the ground truth VGP pairs in the training split as positive  instances. During training, we constructed mini-batches with 15\% of  positive  instances and 85\% of randomly sampled negative instances.
We used Adam for optimization with a mini-batch size of 300 and weight decay of 0.0001. The learning rate was initialized to 0.01, which was halved at every epoch. We used sigmoid cross entropy loss. We terminated training after 5 epochs, where we observed the loss converged on the validation split. For the entity feature vectors, we compared three different settings described above namely: WEA, FV, and FV+CCA. 
\item SNN+image: this setting is for our proposed supervised NN-based method described in Section \ref{sec:nn}. We again compared the three different entity feature vectors. We used VGG-16 \cite{DBLP:journals/corr/SimonyanZ14a} for the image features. The model was trained with the same configuration as the SNN setting.
\item Ensemble: the ensemble of the SNN and SNN+image models that takes the average similarity given by both models. The motivation of this setting is to complement these two models to each other.
\end{itemize}

\subsection{Results}
\begin{table*}[t]
\small
\begin{center}
\begin{tabular}{l|c|c|c|c}\hline
\bf  & \bf ARI & \bf Precision &\bf Recall &\bf F-score \\
\bf Method & \bf  all / single / multi & \bf   all / single / multi  &\bf  all / single / multi  &\bf   all / single / multi \\ \hline 
PL & 43.23 / 45.92 / 46.35 & 59.32 / 51.53 / 62.86 & 63.12 / 47.99 / 74.14 & 61.16 / 49.70 / 68.04  \\ \hline 
TP & 37.61 / 50.32 / 36.79 & 66.23	/ 63.20 /	{\bf 82.17} &	64.20 /	66.10 /	56.31 &	65.20 /	64.62 /	66.83 \\
WEA & 49.82 / 47.16 / 49.58 & 61.51	/ 46.11 /	62.84 &	71.29 /	67.93 /	78.47 &	66.04 /	54.93 /	69.79 \\
FV  & 39.85 / 43.55 / 41.26 & 63.79	/ 40.84 /	67.51 &	60.87 /	35.41 /	 77.03 &	62.30 /	37.94 /	71.96 \\
FV+CCA & 54.91 / 51.01 / 49.30 & 64.79 / 55.79 / 68.24 &	82.20 / 75.83 / 84.98 & 72.46 / 64.28 / 75.69 \\ \hline
SNN (WEA)  &  60.23 / 55.06 / 53.26 & 77.86	/ 83.66 / 74.50 &	84.58 /	75.16 /	{\bf 88.96} &	81.08 /	79.18 /	81.09 \\
SNN+image (WEA)  & 60.55 / 55.42 / {\bf 55.82} &  79.47 / 81.01 /	77.26 &	84.56 /	79.35 /	87.06 &	81.94 /	80.17 /	81.86 \\
Ensemble (WEA)  & 60.65 / 54.92 / 54.56 &  80.65	/ 78.68 /	77.38 &	84.79 /	83.14 /	88.85 &	82.67 /	80.85 /	82.72 \\
SNN (FV) & 48.13 / 45.97 / 47.22 &  64.21	/ 45.92 / 66.40 &	65.93 /	50.89 /	76.51 &	65.06 /	48.28 /	71.10 \\
SNN+image (FV) & 47.90 / 47.39 / 48.31  &  63.49	 / 52.62 /	66.86 &	68.20 /	55.62 /	78.01 &	65.76 /	54.08 /	72.01 \\
Ensemble (FV)  & 49.82 / 48.16 / 48.34 &  65.48	/ 54.87 /	70.51 &	71.43 /	56.24 /	76.54 &	68.33 /	55.55 /	73.40 \\
SNN (FV+CCA) &   60.56 / {\bf 56.35} / 54.06  &  {\bf 83.11} / {\bf 85.19} / 77.44 &	82.13 /	79.30 /	87.69 &	82.62 /	82.14 /	82.25 \\
SNN+image (FV+CCA) & 61.17 / 54.86 / 54.14 &  82.51	/ 84.52 / 80.28 &	84.19 /	81.85 /	86.82 &	83.34 /	83.16 /	83.43 \\
Ensemble (FV+CCA)  & {\bf 62.35} / 54.98 / 54.84 &  82.71 / 84.10 /	80.91 &	{\bf 85.67} / {\bf 83.50} /	87.06 &	{\bf 84.16} / {\bf 83.80} / {\bf 83.87} \\
\hline 
\end{tabular}
\end{center}
\vspace{-3mm}
\caption{\label{table:new_structure} VGP extraction results (``all'' evaluates on all entities, ``single'' and ``multi'' only evaluate on entities consist of one single token and multiple tokens after removing stop words, respectively; the methods above and below the double line are unsupervised and supervised, respectively).}
\vspace{-2mm}
\end{table*}

Table \ref{table:new_structure} shows the results of all the different methods. We report the performance based on the entity types to better understand the performance difference of each method, i.e., ``all'' evaluates on all entities, whereas ``single'' and ``multi'' only evaluate on entities with one single token and multiple tokens, respectively, after removing stop words. For the unsupervised methods, we can see that PL does not show good performance. This is due to the low performance of phrase localization.\footnote{Although \cite{Plummer_2017_ICCV} is the current state-of-the-art for phrase localization, the accuracy is only 55.85\%.} TP shows a fairly high F-score, but a very low ARI score. The reason for this is that the translation probabilities are computed based on word alignment, leading to a similarity score of 0 to the unrelated entity pairs, which is not suitable for affinity propagation. WEA shows relatively good performance that is better than FV. This is because 45.84\% of the entities in our task are single word type after removing the stop words, and converting the low dimensional word embedding to high dimensional and sparse Fisher vectors is harmful for these single word entity pairs. However, for the performance of entities containing multiple words, the Fisher vector is better than word embedding average in the perspective of F-score. FV+CCA significantly outperforms FV. This is because it uses visual information in the training split that transforms the entity vectors and visual vectors into the semantic space that is helpful for detecting VGPs. 

Regarding the supervised methods, NN-based methods using any entity feature vectors outperforms the methods that uses them in an unsupervised way. The reason for this is that it directly uses the paraphrase supervision in the training split, while the unsupervised methods do not. Using entity representation with better ARI and F-score for the SNN method can achieve better results. 
Our proposed method (SNN+image) that uses both textual and visual features shows better performance compared to SNN that uses textual features only, indicating that the usage of visual features is helpful for our VGP extraction task.\footnote{The difference between SNN (FV+CCA) and SNN+image (FV+CCA) is whether using visual features for iParaphrasing explicitly or not. SNN (FV+CCA) uses image region features for learning entity features, but it does not use visual features explicitly for iParaphrasing.} However, the performance improvements are not very large. We discuss the reason for this in detail in Section \ref{sec:discussion_image}. The ensemble of SNN and SNN+image further improves the performance, which means that these two models complement each other.


\subsection{Discussion}
\label{sec:discussion}
\subsubsection{Neural Network w/ and w/o Images}
\label{sec:discussion_image}
\begin{figure*}[t]
  \begin{subfigure}[b]{0.5\hsize}
    \includegraphics[width=\hsize]{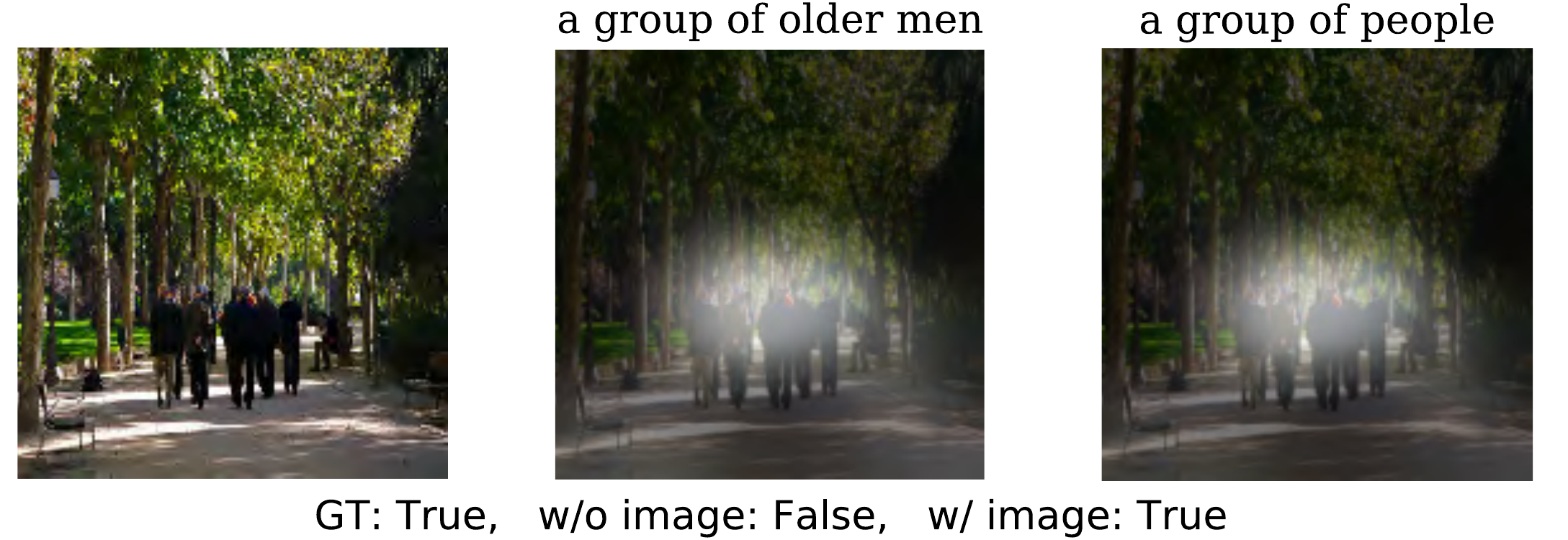}
    \caption{An improved example of people-related paraphrases.}
    \label{fig:image1_improved}
  \end{subfigure}
  \hfill
  \begin{subfigure}[b]{0.5\hsize}
    \includegraphics[width=\hsize]{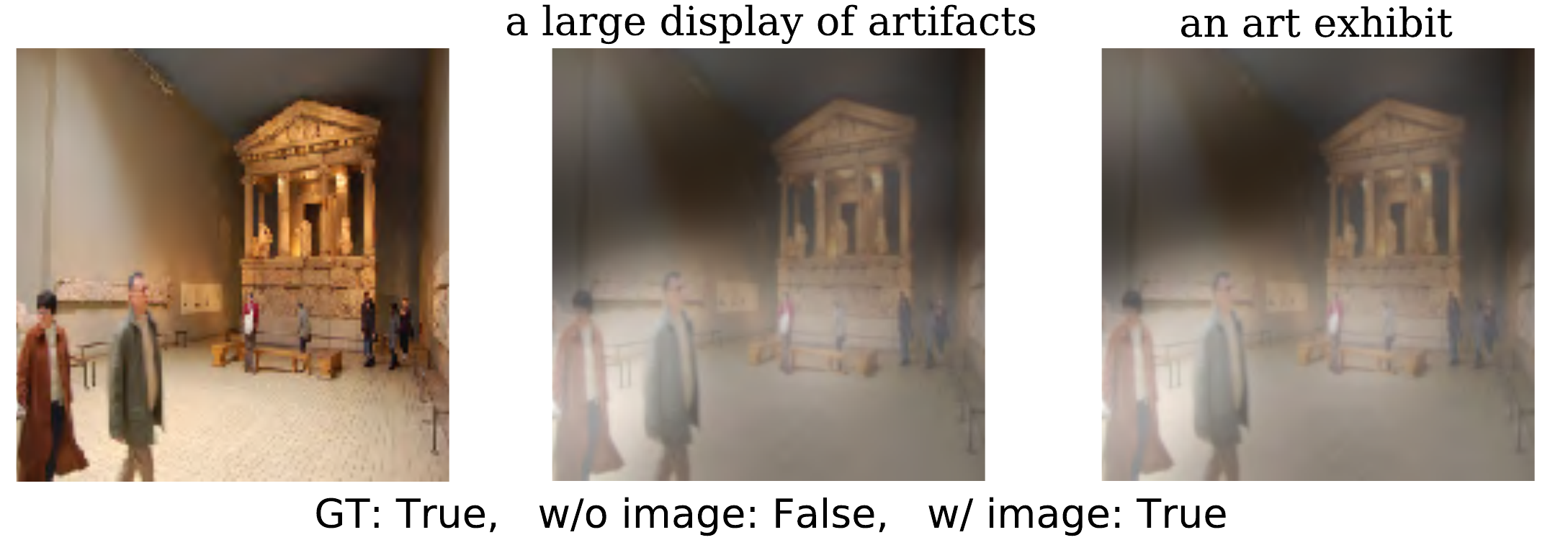}
    \caption{An improved example of scene-related paraphrases.}
    \label{fig:image2_improved}
  \end{subfigure}
  \begin{subfigure}[b]{0.5\hsize}
    \includegraphics[width=\hsize]{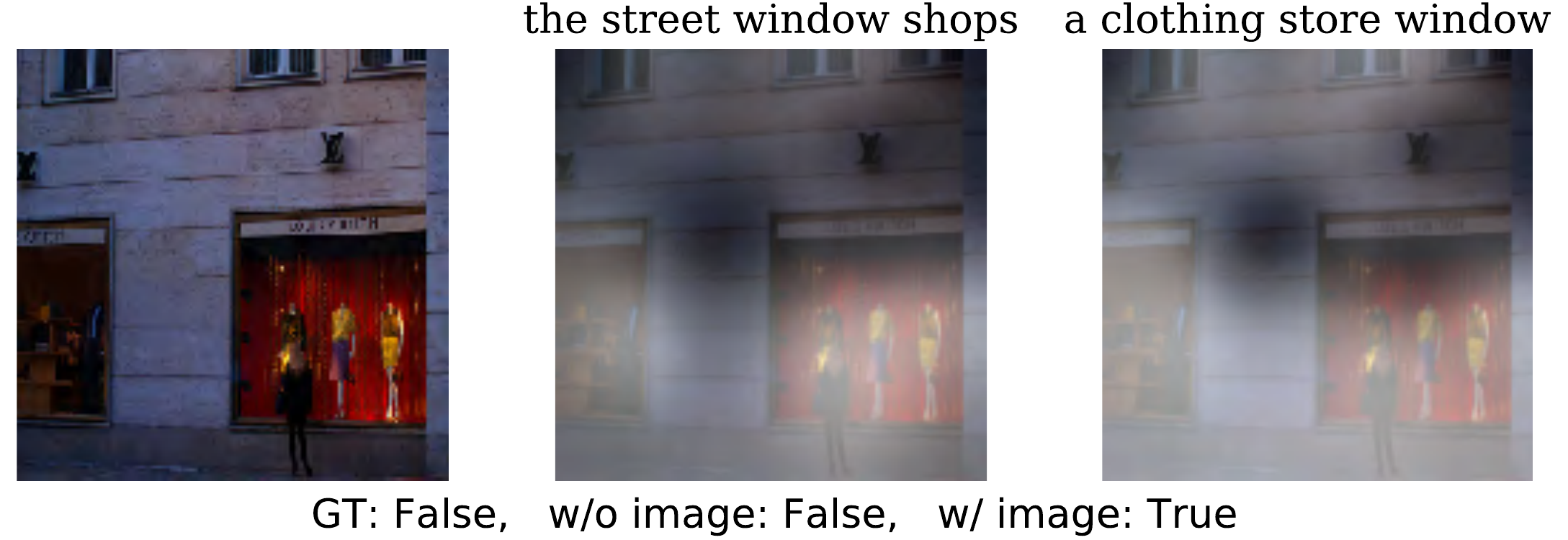}
    \caption{A worsened example of scene-related paraphrases.}
    \label{fig:image1_worsened}
  \end{subfigure}
  \begin{subfigure}[b]{0.5\hsize}
    \includegraphics[width=\hsize]{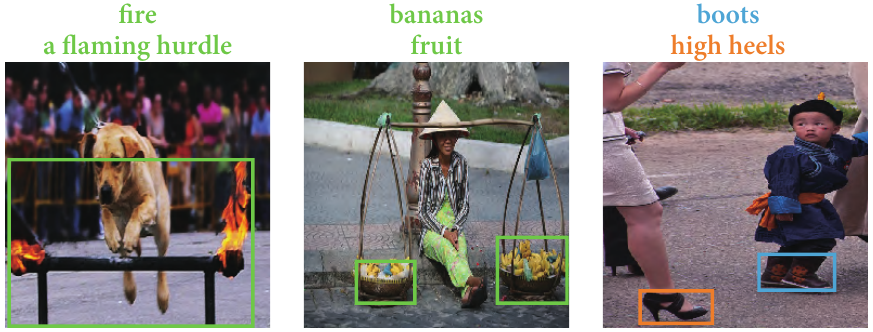}
    \caption{Failed examples.}
    \label{fig:failed_eg}
  \end{subfigure}
  \vspace{-4mm}
  \caption{Examples comparing SNN (FV+CCA) with SNN+image (FV+CCA) ((a), (b), and (c); the leftmost images are the original ones, the images in the middle and on the right show attention of the entity pairs on the images, the degree of whiteness indicates the strength of attention, the identification results are shown under the images), and failed examples (d).}
  \label{fig:image_examples}
  \vspace{-2mm}
\end{figure*}

We compared the SNN and SNN+image results, and found that image attention is helpful for identifying people-related paraphrases in about 50\% cases, which are difficult to be determined based on the textual information only. In addition, the attention for these people-related paraphrases are well learned. We believe the reason for this is that many entities in the training split are people-related and thus they are well modeled. Figure \ref{fig:image1_improved} shows such an example, where the SNN (FV+CCA) model fails to identify these two entities ``a group of order men'' and ``a group of people'' as VGPs due to the diverse textual descriptions of the the same visual concept. Our proposed NN+image (FV+CCA) model correctly identifies these VGPs by paying attention to the image region of people in the image. In about 30\% cases, the visual information is also helpful for the identification of other types of paraphrases, although the attention is not accurate. Figure \ref{fig:image2_improved} shows an example, where the SNN (FV+CCA) model could not identify two entities ``a large display of artifacts" and ``an art exhibit'' as VGPs. 

In about 20\% cases, visual information could bring negative effects for paraphrase identification. 
Figure \ref{fig:image1_worsened} shows an example that ``the street window shops'' and ``a clothing store window'' are mistakenly judged as a paraphrase after using the image information while using  textual information judges correctly. Although, the attention for these entities refer to the same visual concept in the image, the entities actually refer to different concepts (i.e., ``shop'' and ``window'').


\subsubsection{Failed Examples}
\label{sec:failed}
Even the best method, namely Ensemble (FV+CCA), only achieves a ARI of $62.35$ and a F-score of $84.16$.
We found that most false negative examples are sparse entity pairs that describe a image region in an image in a very diverse way, for example ``fire'' and ``a flaming hurdle'' (Figure \ref{fig:failed_eg} (left)), ``bananas'' and ``fruit'' (Figure \ref{fig:failed_eg} (middle)). These pairs are difficult not only for using textual features, but also for using image attentions. 
Most false positive examples are produced by the noisy phrase embedding method. For example, ``boots'' and ``high heels'' referring to the shoes on a boy and a lady, respectively, are identified as a paraphrase pair because of their closeness in the embedding space (Figure \ref{fig:failed_eg} (right)).
Some of the false positive examples are caused by the noise introduced by the wrong attention in an image. For example, ``a green snowman'' and ``his new toy'' are attended to the similar image regions.









\section{Conclusion}
In this paper, we proposed iParaphrasing: a novel task to extract VGPs describing the same visual concept in an image. We not only applied various existing techniques for this task, but also proposed a NN-based method that uses both the textual and visual information to model the similarity between the VGPs. Experiments on the Flickr30k entities dataset showed that we achieved good performance.

For future work, we plan to study a multi task method for both VGP extraction and phrase localization to further improve the performance. We worked on the Flickr30k entities dataset, where noun phrases are given and VGP supervision is available, extracting VGP in an end-to-end manner without supervision in other datasets such as the Microsoft COCO caption dataset \cite{DBLP:journals/corr/LinMBHPRDZ14} is a more
realistic scenario and could be more interesting. Extracting other types of paraphrases (e.g., prepositional and verb paraphrases) is another possible extension, which requires a much deeper understanding of the relation between phrases and image regions. 
We also plan to apply the VGPs for CV and NLP multimodal tasks, such as VQA.

\subsubsection*{Acknowledgement}
This work was supported by ACT-I, JST and JSPS KAKENHI No.~18H03264. We are very appreciated to Prof. Kumiyo Nakakoji, Prof. Yuki Arase and Prof. Sadao Kurohashi for the helpful discussion of this paper. We also thank
the anonymous reviewers for their insightful comments.

\bibliographystyle{acl}
\bibliography{coling2018}

\end{document}